\documentclass{article}

\usepackage[final]{neurips_2026}
\makeatletter
\providecommand{\@trackname}{}
\makeatother

\usepackage{amsmath, amssymb}
\usepackage{graphicx}
\usepackage{booktabs}
\usepackage{hyperref}
\usepackage{caption}
\usepackage{longtable}
\usepackage{geometry}
\usepackage{enumitem}
\usepackage{makecell}
\usepackage{listings} 
\newcommand{\keywords}[1]{%
  \noindent\textbf{Keywords:} #1
}

\graphicspath{{figures/}}

\title{Placement Is Free, Composition Is Not: The Latin Square as a Provably-Balanced Construction for Heterogeneous Sequence-Mixer Stacks}
\author{
Taebong Kim \quad
Youngsik Hong \quad
Minsik Kim \quad
Sunyoung Choi \\
Jaewon Jang \quad
Minseo Kim \\
VIDRAFT AI Research $\cdot$ QuantumOS, Seoul, Republic of Korea \\
\texttt{arxivgpt@gmail.com}
}

\begin{document}
\maketitle

\begin{abstract}

Since GPT, most transformers have repeated the \textbf{same attention mechanism at every layer}. 
Yet this design is largely a convention rather than a tested conclusion. When multiple sequence mixers are combined in one stack, 
improvements may arise from the choice of mechanisms, their placement, or both, making causal attribution difficult.

We introduce \textbf{Aether-7B-5Attn}, a 6.59B-parameter mixture-of-experts model ($\approx$2.98B active) whose 49 layers contain seven sequence-mixing mechanisms arranged 
as a \textbf{$7\times7$ Latin square}. Because each mechanism appears exactly once in every row and column, 
the construction guarantees balanced exposure across depth while equalizing marginal position across mechanisms.

To evaluate this principle, we build a parameter-matched proxy with four mechanisms arranged in a $4\times4$ Latin square over sixteen layers, 
matched to \textbf{700.9M parameters} and trained with \textbf{eight seeds per arm}. The results reveal a clear \textbf{three-way dissociation}. 
Rearranging a distributed heterogeneous stack into a balanced periodic cycle changes validation loss by only \textbf{0.16\%}, 
indicating that exact placement is largely irrelevant. Clustering the same mechanisms into contiguous depth bands incurs a \textbf{0.59\%} penalty, 
while replacing the heterogeneous stack with a homogeneous one incurs a \textbf{1.68\%} penalty.

These findings show that performance depends primarily on \textbf{balanced composition distributed across depth}, 
not on any specific permutation. We confirm the composition result at \textbf{$2.16\times$ larger scale} (1.514B parameters), 
where the homogeneous-stack penalty increases to \textbf{2.63\%} and the penalty from removing the SSM-family mechanism reaches \textbf{3.20\%}.

We additionally report per-mechanism cost profiles, English and Korean evaluations, 
and a causal-safety audit of all 49 layers. We release model weights, training-data recipes, training code, logs, and architecture source code.

\end{abstract}


\section{Introduction}

Transformer language models have traditionally employed a homogeneous architectural design in which the same sequence-mixing mechanism is repeated throughout depth. 
From the original Transformer \citep{Vaswani2017AttentionIA} to modern large language models including GPT, GPT-2, GPT-3, PaLM, Chinchilla, 
and LLaMA \citep{radford2018improving,radford2019language,brown2020gpt3,chowdhery2022palm,hoffmann2022training,touvron2023llama}, 
attention layers are typically replicated with an identical computational structure, differing only in learned parameters. 
This design has achieved remarkable empirical success and has therefore become the default architectural paradigm.

At the same time, a growing body of work has introduced alternative sequence-mixing mechanisms, including state-space models, recurrent operators, linear-attention variants, 
localized attention schemes, and hybrid architectures. These mechanisms differ substantially in their inductive biases, computational complexity, memory requirements, 
and long-context behavior. Consequently, there is no obvious reason to expect a single mechanism to be optimal at every depth of a deep network.

Recent work has explored a diverse set of sequence-mixing architectures. Efficient attention variants include Longformer, BigBird, Linformer, Reformer, Performer, Routing Transformer, 
Nystr\"omformer, FlashAttention, FlashAttention-2, Differential Transformer, and Multi-Head Latent Attention
 \citep{beltagy2020longformer,zaheer2020bigbird,wang2020linformer,kitaev2020reformer,choromanski2021performer,roy2021routing,xiong2021nystromformer,dao2022flashattention,dao2023flashattention2,ye2024differential,deepseek2024mla}. 
Beyond attention, state-space models have evolved from HiPPO and S4 through S4D and S5 to Mamba and Mamba-2 \citep{gu2021s4,gu2022s4d,smith2023s5,gu2023mamba,dao2024mamba2}, 
while recurrent and retention-based approaches include Transformer-XL, Compressive Transformer, RetNet, RWKV, 
and xLSTM \citep{dai2019transformerxl,rae2020compressive,sun2023retnet,peng2023rwkv,beck2024xlstm}. Convolutional approaches such as ConvBERT, Hyena, 
and HyenaDNA provide yet another family of alternatives \citep{jiang2020convbert,poli2023hyena,nguyen2023hyenadna}. 
Hybrid architectures including Memorizing Transformers, Griffin, RecurrentGemma, Jamba, Zamba, 
and Nemotron-H combine multiple sequence-processing mechanisms within a 
single network \citep{wu2022memorizing,de2024griffin,recurrentgemma2024,lieber2024jamba,zamba2024,nemotronh2025}. 
Despite their differences, most existing work focuses on designing new mixers or improving computational efficiency rather than 
studying how heterogeneous mechanisms should be allocated across depth.

This observation motivates heterogeneous architectures that combine multiple sequence-mixing mechanisms within a single model.
However, understanding why such architectures succeed is surprisingly difficult. When several mechanisms are placed in one stack, 
any observed improvement can arise from at least three sources: the set of mechanisms employed, their placement within depth, 
or changes in parameter allocation. Because these factors are typically varied simultaneously, their effects become difficult to disentangle retrospectively. 
Existing heterogeneous architectures generally rely on manually designed layer schedules, making composition and placement effects difficult to separate empirically.

In this work we isolate these factors through a construction based on Latin squares. 
A Latin square arranges $N$ mechanisms across $N^2$ layers such that each mechanism appears exactly once in every row and every column. 
Applied to sequence-mixer design, this construction guarantees balanced exposure across depth while equalizing marginal position across mechanisms. 
The same property makes the resulting architecture a useful experimental instrument: 
composition effects can be studied without introducing depth-concentration biases, even though first-order carryover is not balanced by this construction.

We instantiate this idea in \textbf{Aether-7B-5Attn}, a 6.59B-parameter mixture-of-experts language model ($\approx 2.98$B active parameters) whose 49 layers contain 
seven sequence-mixing mechanisms arranged as a $7\times7$ Latin square. To examine the underlying principle independently of flagship-scale training costs, 
we construct a parameter-matched proxy using four mechanisms arranged on a $4\times4$ Latin square and evaluate multiple placement and composition variants under controlled conditions.

Our experiments reveal a clear three-way dissociation. First, reordering mechanisms within a balanced distributed schedule produces negligible changes in validation loss, 
indicating that permutation is largely irrelevant. Second, concentrating mechanisms into contiguous depth bands degrades performance. 
Third, replacing a heterogeneous stack with a homogeneous one incurs the largest penalty. 
Together, these results indicate that performance is primarily associated with balanced composition distributed throughout depth rather than with any specific ordering of mechanisms.

To our knowledge, this is the first work to study heterogeneous sequence-mixer allocation as a controlled experimental variable and to provide 
a search-free construction that guarantees balanced depth-wise distribution by design.

The contributions of this paper are as follows:
\begin{enumerate}
\item We introduce the Latin square as a provably balanced, search-free construction for heterogeneous sequence-mixer architectures.
\item We release Aether-7B-5Attn, a 6.59B-parameter mixture-of-experts model employing seven sequence-mixing mechanisms across a $7\times7$ Latin-square layout.
\item We present a parameter-matched, multi-seed ablation that separates permutation, distribution, and composition effects, 
demonstrating that composition and depth distribution dominate placement.
\item We provide the first detailed per-mechanism cost profile for a heterogeneous sequence-mixer stack across multiple context lengths.
\item We release model weights, architecture source code, training recipes, logs, and evaluation artifacts to facilitate reproducibility.
\end{enumerate}

\section{Related Work}

\paragraph{Sequence-mixing mechanisms.}

The Transformer established self-attention as the dominant sequence-mixing mechanism for modern language models \citep{Vaswani2017AttentionIA}. 
The same basic design has subsequently been adopted by BERT, GPT, GPT-2, GPT-3, PaLM, Chinchilla, 
and LLaMA \citep{devlin-etal-2019-bert,radford2018improving,radford2019language,brown2020gpt3,chowdhery2022palm,hoffmann2022training,touvron2023llama}. 
Despite their differences in scale and training methodology, these models retain a largely homogeneous architecture in which the same mixer is repeated throughout depth.

A large literature has sought to improve or replace standard attention. Efficient attention variants include Longformer, BigBird, Linformer, Reformer, Performer, 
Routing Transformer, Nystr\"omformer, FlashAttention, FlashAttention-2, Differential Transformer, 
and Multi-Head Latent Attention (MLA) \citep{beltagy2020longformer,zaheer2020bigbird,wang2020linformer,kitaev2020reformer,choromanski2021performer,roy2021routing,xiong2021nystromformer,dao2022flashattention,dao2023flashattention2,ye2024differential,deepseek2024mla}. These methods primarily target computational efficiency, memory reduction, or improved long-context behavior while preserving the attention paradigm.

Beyond attention, alternative sequence-processing mechanisms have emerged. State-space models include HiPPO, S4, S4D, S5, Mamba, 
and Mamba-2 \citep{gu2021s4,gu2022s4d,smith2023s5,gu2023mamba,dao2024mamba2}, while retention-based and recurrent approaches include RetNet, RWKV, 
and xLSTM \citep{sun2023retnet,peng2023rwkv,beck2024xlstm}. Convolutional approaches such as ConvBERT, Hyena, and HyenaDNA provide additional alternatives 
to attention-based sequence modeling \citep{jiang2020convbert,poli2023hyena,nguyen2023hyenadna}. Collectively, these works demonstrate that effective sequence modeling can 
arise from a diverse set of inductive biases rather than from self-attention alone.

\paragraph{Hybrid sequence architectures.}

Recent work has increasingly combined heterogeneous sequence-processing mechanisms within a single architecture. Early examples include Transformer-XL 
and Compressive Transformer, which augment attention with recurrent memory structures \citep{dai2019transformerxl,rae2020compressive}. 
More recent architectures such as Memorizing Transformers, Griffin, RecurrentGemma, Jamba, Zamba, and Nemotron-H combine attention, recurrent operators, 
state-space models, or external memory mechanisms within a unified design \citep{wu2022memorizing,de2024griffin,recurrentgemma2024,lieber2024jamba,zamba2024,nemotronh2025}. 
These models demonstrate that heterogeneous architectures can achieve favorable trade-offs between expressiveness, efficiency, and long-context capability.

However, existing hybrid architectures generally treat layer allocation as an implementation detail rather than as an object of study. 
Different mechanisms are typically assigned to depth according to architecture-specific design choices, making it difficult to determine 
whether performance gains arise from composition itself, from the placement of those mechanisms, or from both simultaneously.

\paragraph{Position of this work.}

Our work differs from prior heterogeneous architectures in two respects. First, we treat mixer allocation across depth as the primary experimental variable 
rather than as a fixed architectural choice. Second, we introduce a Latin-square construction that guarantees balanced depth-wise exposure of every mechanism by construction 
while eliminating placement asymmetries without architecture search. This enables composition and placement effects to be disentangled under parameter-matched conditions. 
To our knowledge, no previous work has used a Latin-square design to study heterogeneous sequence-mixer allocation in a controlled experimental setting.

\section{Design}

\subsection{Mechanisms carry different inductive biases}

Table~\ref{tab:mechanisms} summarizes the five base sequence-mixing mechanisms used throughout Aether.
Each mechanism is optimized for a different objective: full attention provides exact pairwise interactions,
sliding attention trades global reach for efficient locality, differential attention suppresses common-mode noise,
linear-recurrent mixing offers scalable long-context processing, and NSA combines multiple pathways to obtain
cheap long-range access. None dominates across accuracy, efficiency, memory, and context length simultaneously.
Consequently, repeating a single mechanism throughout the network also repeats its weaknesses throughout the network.
This observation motivates heterogeneous composition: rather than selecting one mixer and applying it at every depth,
we distribute complementary mechanisms across layers so that their strengths can compensate for one another.

\begin{table}[t]
\centering
\small

\caption{
Five base sequence-mixing mechanisms used in Aether.
Each mechanism provides a distinct inductive bias, computational profile, and failure mode;
the motivation for heterogeneous architectures is that no single mechanism dominates across all dimensions.
}
\label{tab:mechanisms}

\begin{tabular}{l p{6.0cm} p{5.0cm}}
\toprule
Type & Strength & Cost \\
\midrule

\texttt{full}
&
Exact all-pairs interactions; no information loss
&
\textbf{Quadratic} in length
\\

\texttt{sliding}
&
Locality very cheaply, linear in length
&
Blind beyond the window
\\stronger uniformity

\texttt{differential}
&
Subtracts two attention maps, cancelling common-mode noise
&
Halves the head dimension
\\

\texttt{linear}
&
\textbf{Linear-recurrent (Mamba-style) mixing}; linear in length
&
Approximate---no exact pairwise scores
\\

\texttt{nsa}
&
Gates compressed, selected, and sliding branches for cheap long reach
&
Structurally complex
\\

\bottomrule
\end{tabular}

\end{table}

\textbf{Five base structures, seven Latin-square slots.} The table above lists the \textbf{five base attention structures}. 
The flagship's 7$\times$7 Latin square cycles \textbf{seven layer-type slots}: these five, plus two that are \textit{composed} 
from them --- \texttt{compress} (a single NSA branch) and \texttt{hybrid} (\texttt{nsa} + \texttt{differential}). 
So "five" (base structures) and "seven" (Latin-square slots) describe the same stack at two granularities; 
§10 notes the label coarseness. 
We call attention to one member specifically: \textbf{\texttt{linear} is a Mamba-style linear-recurrent mixer --- a state-space-family mechanism --- so the flagship's base set is 
not attention-only.}

\textbf{None of the five dominates.} Each is good at something different and pays a different price --- which means picking one and repeating it 49 times also repeats its weakness 49 times.

\subsection{The Latin square is the construction that guarantees the property that matters}

A Latin square places \textbf{each type exactly once in each row and each column} --- equivalently, 
exactly once in every aligned block of $N$ consecutive layers, and exactly once at each within-block position across the $N$ blocks. 
This is a different balance property from a plain periodic cycle, not a strictly stronger one. 
The periodic cycle fixes the column assignment, but it attains balance in every sliding window of $N$ layers, which a Latin square provably cannot: 
if every length-$N$ sliding window were balanced, comparing two consecutive windows forces the departing symbol to equal the entering one, 
i.e. $x_{i+N} = x_i$, so the schedule would be $N$-periodic.

What the Latin square guarantees instead is that no type concentrates at any depth and no type is pinned to a fixed within-block position. 
No type can concentrate at any depth, at \textbf{any $N$}, with \textbf{no search} 
and a \textbf{proof} rather than a hand-tuned schedule.

Two things follow, and they are the same thing seen twice:

\begin{itemize}
\item \textbf{As engineering}, if the balance-and-distribution property turns out to be what matters (§6 shows it does), then the Latin square is a principled default: 
it delivers that property \textbf{by construction}, at arbitrary depth, with a proof rather than by hand-design and search. 
It is not the \textit{only} schedule that does so --- a plain periodic cycle also keeps every mechanism distributed --- but the Latin square is the \textbf{canonical, 
maximally-uniform} such construction, generalizing to any prime \textit{N} and guaranteeing the property provably rather than incidentally.
\item \textbf{As experiment}, that same uniformity makes placement statistically flat by construction, 
so the question \textit{"what does mechanism composition do?"} becomes answerable without a placement confound.
\end{itemize}

\begin{quote}
\textbf{To our knowledge, no released model places heterogeneous attention in a Latin square.}
\end{quote}
Latin balancing equalizes marginal position frequencies; it does not equalize first-order carryover. 
In the 4x4 proxy, only 7 of the 12 ordered adjacent pairs occur (frequencies 3,3,3,3,1,1,1), so adjacent-mixer transition structure is not balanced by this construction.

\subsection{The prediction, and why placement-invariance is the point}

If balance-and-distribution is the operative property, three things should hold: (a) re-permuting within a balanced, 
distributed schedule should \textit{not} change quality --- permutation is free; (b) \textbf{breaking distribution} by clustering the same mechanisms should \textit{hurt}; 
(c) \textbf{breaking balance} by homogenizing should hurt \textit{more}. 65 tests all three, and all three hold.

We state the logic plainly because it inverts a natural objection. \textit{"If placement doesn't matter, why the Latin square?"} --- Because placement-invariance is 
exactly the evidence that the \textbf{property}, not the permutation, carries the benefit; and once that is true, the construction that \textit{guarantees} the property without search is 
the right one to ship. Placement-invariance is not the Latin square's weakness. It is its justification.

\section{Architecture}

\subsection{Specification}

Table~\ref{tab:specification} summarizes the architecture of Aether-7B-5Attn. 
The model combines a mixture-of-experts backbone with a heterogeneous sequence-mixer stack arranged as a $7\times7$ Latin square over 49 layers. 
We report the primary architectural hyperparameters, including parameter counts, expert configuration, hidden dimensions, attention configuration, 
vocabulary size, and training context length.

\begin{table}[t]
\centering
\small
\caption{
Architectural specification of Aether-7B-5Attn.
Active parameters denote the average number of parameters participating in each token computation under MoE routing.
}
\label{tab:specification}

\begin{tabular}{ll}
\toprule
Item & Value \\
\midrule
Total parameters & \textbf{6.59B} \\
Active parameters & \textbf{$\approx$2.98B} per token \\
Layers & \textbf{49 ($7\times7$ Latin square)} \\
Experts & 25, top-7 routing, 1 shared \\
Expert intermediate & 640 \\
Hidden / intermediate & 2048 / 6144 \\
Heads / KV heads / head\_dim & 16 / 4 / 128 \\
Vocabulary & 151,936 \\
Training context & 4096 \\
Dtype & bfloat16 \\
\bottomrule
\end{tabular}
\end{table}

\subsection{Measured cost profile per mechanism}

The sequence-mixing mechanisms used in Aether differ not only in inductive bias but also in computational characteristics. 
To quantify these differences, we benchmark each mechanism as an isolated layer under identical implementation and hardware conditions. 
Because latency and memory consumption are architectural properties rather than properties of particular trained weights, 
these measurements are independently reproducible and provide a direct comparison of the efficiency trade-offs associated with each mechanism.

Table~\ref{tab:cost-profile} reports prefill latency and peak memory usage at context lengths of 2K, 8K, and 32K tokens. 
The objective is not to identify a universally optimal mechanism, but rather to characterize the distinct operating regimes that motivate heterogeneous composition.

\begin{table}[t]
\centering
\small
\caption{
Per-mechanism computational cost profile.
Values report prefill latency (ms) and peak memory usage (GB) for an isolated layer at different context lengths.
}
\label{tab:cost-profile}

\begin{tabular}{llll}
\toprule
Type & 2K (ms / GB) & 8K (ms / GB) & 32K (ms / GB) \\
\midrule
\texttt{full} & 0.4 / 0.0 & 1.5 / 0.2 & 13.6 / 0.7 \\
\texttt{differential} & 0.5 / 0.1 & 3.7 / 0.3 & 46.5 / 1.1 \\
\textbf{\texttt{sliding}} & 0.6 / 0.1 & 1.9 / 0.2 & \textbf{7.6 / 0.8} \\
\texttt{nsa} & 1.0 / 0.1 & 3.6 / 0.3 & 26.8 / 3.5 \\
\texttt{hybrid} & 1.5 / 0.1 & 7.7 / 0.3 & 74.7 / 3.5 \\
\bottomrule
\end{tabular}
\end{table}

The measurements confirm that different sequence mixers occupy distinct efficiency regimes. At short contexts, full attention remains competitive, achieving the lowest latency at 2K tokens. 
As context length increases, however, its quadratic scaling becomes increasingly costly. 
In contrast, \texttt{sliding} attention exhibits substantially better scaling behavior and becomes the fastest mechanism at 32K, where it is approximately \textbf{1.8$\times$ faster} 
than \texttt{full}. This behavior is consistent with its locality-constrained design.

The remaining mechanisms illustrate additional trade-offs. \texttt{differential} attention incurs higher computational cost in exchange for enhanced representational capacity, 
while \texttt{nsa} trades efficiency for richer long-range routing behavior. The \texttt{hybrid} mechanism combines \texttt{nsa} and \texttt{differential} components, 
and its latency correspondingly approaches the aggregate cost of its constituents. At the system level, 
the complete Aether model processes a 32K-token context within \textbf{3.46\,GB} of memory.

To our knowledge, no prior work has reported a measured per-mechanism cost profile for a heterogeneous sequence-mixer architecture. 
We therefore include these measurements as a practical complement to the architectural analysis presented throughout this paper.

\section{Training and Release}

\subsection{Data --- specified to the point of reconstruction}

We do not describe the corpus qualitatively. We give the source repository, configuration, token count, and sampling weight for every component, together with the tokenizer 
and document-boundary convention, so that \textbf{the training corpus can be reconstructed byte-for-byte} from the paper alone.

Table~\ref{tab:data} summarizes all components of the training corpus together with their source repositories, configurations, token counts, and licenses. 
By specifying the data mixture at this level of detail, we make the training set reproducible and auditable rather than relying on qualitative dataset descriptions.

\begin{table}[t]
\centering

\caption{
Training corpus composition. We report the source repository,
configuration, token count, sampling weight, and license information
for each component used in the data mixture.
}
\label{tab:data}

\small
\begin{tabular}{p{2.4cm} p{3.0cm} p{2.2cm} p{1.3cm} p{0.9cm} p{1.7cm}}
\toprule
Component & Repository & Config & Tokens & Weight & License \\
\midrule

English web (edu-filtered)
&
\makecell[l]{\texttt{HuggingFaceFW/}\\\texttt{fineweb-edu}}
&
\texttt{sample-100BT}
&
15.000B
&
\textbf{2.0}
&
ODC-By
\\

Synthetic textbook
&
\makecell[l]{\texttt{HuggingFaceTB/}\\\texttt{smollm-corpus}}
&
\texttt{cosmopedia-v2}
&
8.000B
&
\textbf{2.0}
&
ODC-By
\\

Math (filtered)
&
\makecell[l]{\texttt{HuggingFaceTB/}\\\texttt{finemath}}
&
\texttt{finemath-3plus}
&
6.002B
&
\textbf{3.5}
&
ODC-By
\\

Code
&
\makecell[l]{\texttt{OpenCoder-LLM/}\\\texttt{opc-fineweb-}\\\texttt{code-corpus}}
&
default
&
5.001B
&
\textbf{2.5}
&
MIT
\\

Math (web)
&
\makecell[l]{\texttt{open-web-math/}\\\texttt{open-web-math}}
&
default
&
4.004B
&
\textbf{3.5}
&
source repo
\\

\textbf{Korean web}
&
\makecell[l]{\texttt{HAERAE-HUB/}\\\texttt{KOREAN-WEBTEXT}}
&
default
&
\textbf{2.492B}
&
\textbf{2.0}
&
source repo
\\

\textbf{Korean synthetic}
&
\makecell[l]{\texttt{HAERAE-HUB/}\\\texttt{KOREAN-}\\\texttt{SyntheticText-1.5B}}
&
default
&
\textbf{1.650B}
&
\textbf{2.0}
&
source repo
\\

\midrule

Earlier blend$^{\dagger}$
&
(composed of the above)
&
---
&
90B
&
\textbf{1.0}
&
---
\\

\midrule

\textbf{Sum}
&
---
&
---
&
---
&
\textbf{18.5}
&
---
\\

\bottomrule
\end{tabular}

\vspace{2pt}
\begin{minipage}{\linewidth} 
  \raggedright 
  \footnotesize 
  $^{\dagger}$ Sampling probability $\propto$ weight $\times$ token-count (normalized). 
  The 90B ``Earlier blend'' at weight 1.0 therefore dominates the effective mixture ($\approx 47\%$ of sampled tokens); the weights above are raw multipliers, not effective proportions. 
  
  \vspace{2pt} 
  
  $^{\ddagger}$ Not additional data. The 90B ``Earlier blend'' is resampled from the same source corpora and is therefore excluded from the 42.1B unique-token total. 
\end{minipage}

\end{table}

Tokenizer: \texttt{Qwen/Qwen3-14B}; \textbf{EOS 151645} inserted at document boundaries; stored as a flat \texttt{uint32} array. 
Sampling weights are published as executable configuration (weights sum to 18.5), with math weighted 3.5, code 2.5, and Korean 2.0 each.

\textbf{Token accounting.} The source files total \textbf{42.1B unique tokens}. 
The 90B "earlier blend" is \textit{not} new data --- it is a phase-1.5 corpus \textbf{resampled from the same sources}, listed for completeness rather than as an additional 90B. 
Training then consumes \textbf{144.2B token-samples} (§5.2): the unique sources drawn repeatedly under the mixing weights above, 
so the trained count exceeds the unique count \textbf{by design}. 
What the recipe reconstructs byte-for-byte is the \textbf{unique source set}; the 144.2B training stream follows deterministically from that set plus the published weights, 
tokenizer, EOS convention, and seed.

\subsection{Compute}

Table~\ref{tab:compute} summarizes the computational resources, optimization settings, and training schedule used to produce the released checkpoint. 
We report hardware configuration, training duration, throughput, optimization hyperparameters, and token budget to facilitate reproducibility 
and provide a transparent account of the computational requirements associated with training Aether-7B-5Attn.

\begin{table}[t]
\centering
\small
\caption{
Training compute, optimization configuration, and resource utilization
for Aether-7B-5Attn.
}
\label{tab:compute}

\begin{tabular}{ll}
\toprule
Item & Value \\
\midrule
Hardware & \textbf{NVIDIA B200 $\times$ 16} (2-node FSDP) \\
Total window & 2026-05-30 $\rightarrow$ 2026-07-16 (\textbf{$\approx$46 days}) \\
Final stage & 30 days 11 hours --- \textbf{$\approx$11,700 B200-hours} \\
Throughput & $\approx$32,000 tok/s \\
Steps / tokens & 162,000 / \textbf{144.2B} \\
Optimizer & AdamW, $\beta=(0.9,0.95)$, $\varepsilon=10^{-8}$ \\
LR schedule & cosine, $5\times10^{-5}\rightarrow5\times10^{-6}$ \\
Weight decay & 0.1 (transformer layers), 0.0 (embeddings, norms) \\
Embedding LR multiplier / layer-wise decay & 0.1 / 0.97 \\
Grad clip / sequence & 1.0 / 4096 \\
Post-training & annealing (released checkpoint) \\
\bottomrule
\end{tabular}
\end{table}

The final model was trained on 16 NVIDIA B200 GPUs over approximately 46 days and consumed 144.2B token-samples. 
We report the complete optimization configuration because training dynamics can substantially influence architectural comparisons. 
Together with the released training logs and intermediate checkpoints, these specifications allow the reported results to be interpreted within their full computational context 
and facilitate independent reproduction of the training procedure.

\subsection{What is released}

Reproducibility is a central objective of this work. Rather than releasing only model weights, 
we provide the artifacts required to reconstruct, inspect, and independently evaluate the reported results. 
Table~\ref{tab:release} summarizes the resources released together with Aether-7B-5Attn, including model checkpoints, source code, training data specifications, and evaluation utilities.

\begin{table}[t]
\centering
\small
\caption{
Artifacts released with Aether-7B-5Attn. The release includes model checkpoints, source code, training recipes, 
and supporting resources intended to facilitate independent verification and reproduction.
}
\label{tab:release}

\begin{tabular}{ll}
\toprule
Item & Released \\
\midrule
Weights (annealed base) & [OK] \\
Full architecture source & [OK] \\
Training-data recipe (sources, configs, token counts, weights, tokenizer, EOS) & [OK] \\
Tokenization script & [OK] \\
Training code and launch scripts & [OK] \\
All hyperparameters & [OK] \\
\textbf{Complete training log} & [OK] \\
Evaluation code & [OK] \\
\textbf{Intermediate checkpoints} (steps 110k / 115k / 162k) & [OK] \\
\bottomrule
\end{tabular}
\end{table}

Weights and source code are released under the \textbf{Apache-2.0} license, while the training corpus inherits the licensing terms of its respective source repositories.

Two components deserve particular emphasis. The \textbf{complete training log} allows readers to verify that the reported optimization trajectory 
and loss curves are consistent with the claims presented in this paper, rather than relying solely on summary statistics. 
The released \textbf{intermediate checkpoints} make it possible to study the training dynamics directly, including the evolution of architectural effects throughout optimization 
and independent re-evaluation of intermediate model states.

Together, these artifacts are intended to support not only reproduction of the final model but also inspection of the training process itself. 
To our knowledge, Aether is the first Korean foundation-model release to provide both a complete training-data recipe and the corresponding training code alongside the model weights.

\section{Ablation --- placement versus composition}
\begin{figure}[t]
\centering
\includegraphics[width=0.82\linewidth]{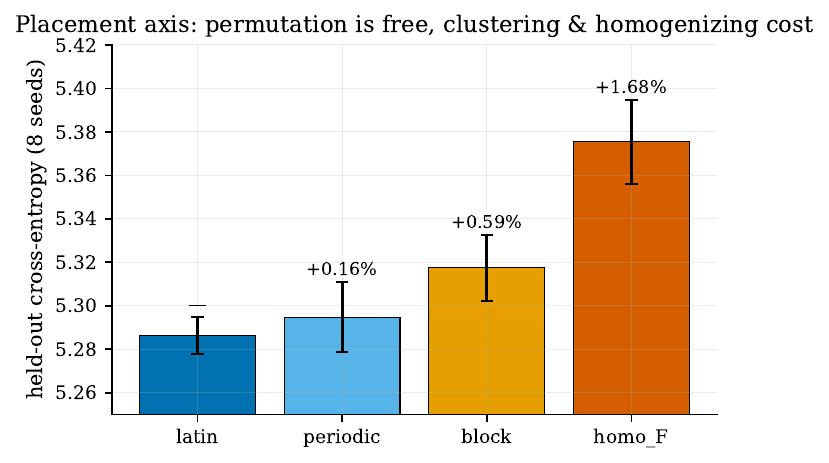}
\caption{Placement axis (8 seeds): re-permuting is free (latin$\approx$periodic), clustering costs 0.59\%, homogenizing costs 1.68\%.}
\label{fig:diss}
\end{figure}
\begin{figure}[t]
\centering
\includegraphics[width=0.82\linewidth]{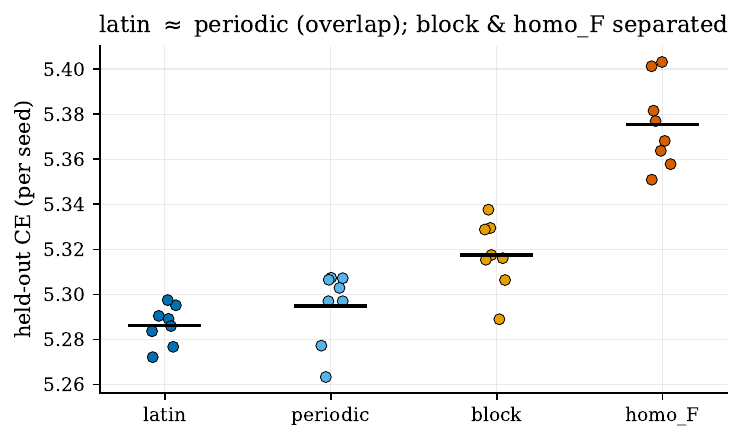}
\caption{Per-seed cross-entropy: latin and periodic overlap; block and homo\_F are cleanly separated.}
\label{fig:scatter}
\end{figure}
\begin{figure}[t]
\centering
\includegraphics[width=0.82\linewidth]{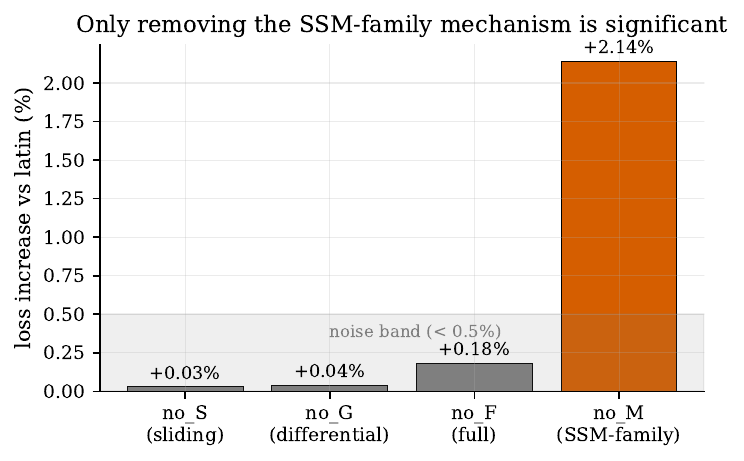}
\caption{Leave-one-out: only removing the SSM-family mechanism is significant.}
\label{fig:loo}
\end{figure}
\begin{figure}[t]
\centering
\includegraphics[width=0.82\linewidth]{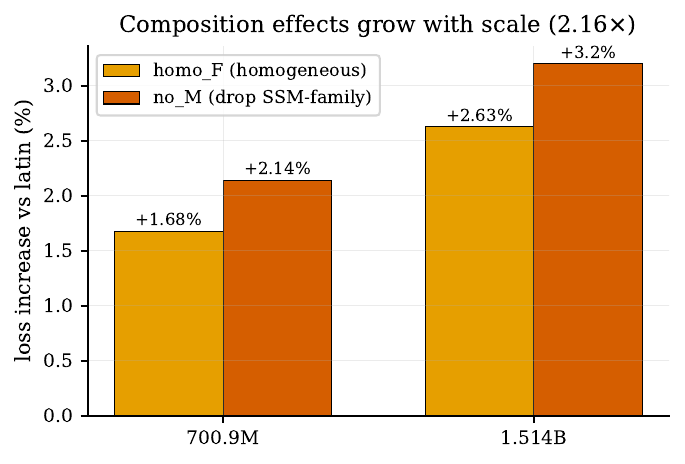}
\caption{Composition effects grow at 2.16$\times$ scale (700.9M$\to$1.514B).}
\label{fig:scale}
\end{figure}
\begin{figure}[t]
\centering
\includegraphics[width=0.82\linewidth]{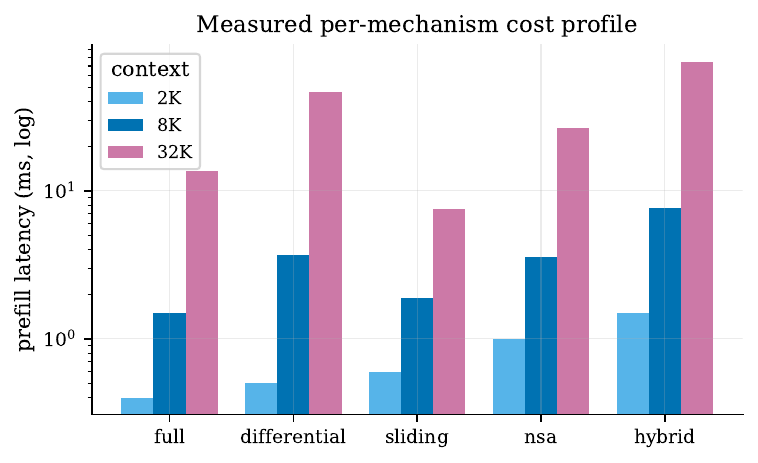}
\caption{Measured per-mechanism prefill cost at 2K/8K/32K.}
\label{fig:cost}
\end{figure}

\subsection{Setup}

The flagship stacks \textbf{seven} sequence mixers on a 7$\times$7 Latin square over 49 layers: 
\texttt{full}, \texttt{sliding}, \texttt{differential}, \textbf{\texttt{linear} (a Mamba-style linear-recurrent mixer)}, \texttt{nsa}, \texttt{compress}, 
and \texttt{hybrid}. The ablation runs in a \textbf{parameter-matched proxy} built from four of these families: 
\texttt{F} (full attention), \texttt{S} (sliding-window attention), \texttt{G} (\textbf{differential attention}), 
and \texttt{M} (\textbf{Mamba-2}, standing in for the flagship's \texttt{linear} as the linear-recurrent / state-space-family member) --- over \textbf{sixteen layers}, 
i.e. a \textbf{4$\times$4 Latin square}, matched to \textbf{700.9M $\pm$0.004\%} across every arm via the feed-forward width. 
The proxy thus shares \textbf{three mechanisms exactly} with the flagship (\texttt{full}, \texttt{sliding}, \texttt{differential}) 
and \textbf{one by family} (Mamba-2 $\leftrightarrow$ \texttt{linear}); it does not include the flagship's \texttt{nsa}, \texttt{hybrid}, 
or \texttt{compress}, which is the coverage limit §10 states. Identical data, optimizer, steps (1,500), and sequence length (1,024). 
Metric: held-out cross-entropy; \textbf{eight seeds} per placement arm, four per composition arm.

Table~\ref{tab:placement-axis} summarizes the placement-axis configurations used in the ablation. 
All three arms contain the same \texttt{4+4+4+4} mechanism multiset and are parameter-matched; the only experimental variable is how those mechanisms are distributed across depth.

\begin{table}[t] 
\centering 
\small 
\caption{ Placement-axis ablation. All arms contain the identical 
\texttt{4+4+4+4} mechanism multiset and identical parameter counts; 
only the depth-wise arrangement of mechanisms differs. 
The Latin-square and periodic schedules maintain distributed exposure, whereas the block schedule concentrates each mechanism within a single depth band. } 
\label{tab:placement-axis} 
\begin{tabular}{lll} 
\toprule Arm & Layer sequence & Property \\
\midrule \texttt{latin} & \texttt{SGMF GMFS MFSG FSGM} & balanced \textbf{and} column-uniform (Latin square) \\
\texttt{periodic} & \texttt{SGMF SGMF SGMF SGMF} & balanced per window, fixed columns (plain cycle) \\
\texttt{block} & \texttt{SSSS GGGG MMMM FFFF} & balanced counts but each type \textbf{confined to a band} \\
\bottomrule 
\end{tabular} 
\end{table}

\texttt{latin} and \texttt{periodic} both distribute every mechanism across depth rather than confining it to a band, but in different senses: 
periodic contains every mechanism in every four-layer sliding window, whereas latin contains every mechanism in every aligned four-layer block and 
at every within-block position across blocks. block confines each mechanism to a single depth band (concentrated).

\textbf{Composition axis} --- one mechanism removed, remaining three cycled (e.g. \texttt{no\_M} = \texttt{SGF…}), re-matched to 700.9M:
\texttt{no\_M} (drop Mamba-2) $\cdot$ \texttt{no\_G} (drop differential) $\cdot$ \texttt{no\_S} (drop sliding) $\cdot$ \texttt{no\_F} (drop full)

\textbf{Reference}: \texttt{homo\_F}, a parameter-matched homogeneous stack (\texttt{F} at every layer) --- the limit of concentration, where three of four mechanisms are absent entirely.

\subsection{Results}

Table~\ref{tab:placement-results} reports the principal placement-axis results together with a parameter-matched homogeneous reference. 
All values are averaged over eight random seeds and correspond to the final experiment confirmed on 2026-07-23. 
Differences are evaluated relative to the Latin-square baseline using the pre-registered decision rule described below.

\begin{table}[t]
\centering
\small
\caption{
Placement-axis results and homogeneous reference (eight seeds per arm; confirmed 2026-07-23).
The Latin, periodic, and block arms share identical mechanism composition and parameter counts, differing only in depth-wise arrangement. The homogeneous reference (\texttt{homo\_F}) represents the limiting case in which a single mechanism occupies every layer.
}
\label{tab:placement-results}

\begin{tabular}{llllll}
\toprule
Arm & Mean CE & SD & $\Delta$ vs.\ \texttt{latin} & $2\cdot$pooled\_SD & Verdict \\
\midrule

\texttt{latin}
& \textbf{5.28639}
& 0.00867
& ---
& ---
& Reference \\

\texttt{periodic}
& 5.29484
& 0.01611
& +0.16\% ($\Delta = 0.00845$)
& 0.02588
& \textbf{Null} (within noise) \\

\texttt{block}
& 5.31754
& 0.01520
& +0.59\% ($\Delta = 0.03115$)
& 0.02475
& \textbf{Real effect} ($2.5\times$ pooled SD) \\

\texttt{homo\_F}
& 5.37540
& 0.01917
& +1.68\% ($\Delta = 0.08901$)
& 0.02976
& \textbf{Real effect} \\

\bottomrule
\end{tabular}
\end{table}

Table~\ref{tab:composition-results} shows that removing attention-based mechanisms produces changes that remain within run-to-run variability, 
whereas removing the Mamba-2 state-space component yields a statistically meaningful degradation, making it the only ablation with a clear effect.

\begin{table}[t]
\centering
\small

\caption{
Composition-axis leave-one-out ablation (four seeds per arm; confirmed 2026-07-23).
Each row removes one mechanism from the heterogeneous stack while re-matching parameters and cycling the remaining three mechanisms throughout depth.
Results are reported relative to the \texttt{latin} baseline.
}
\label{tab:composition-results}

\begin{tabular}{llllll}
\toprule
Arm (removed) & Mean CE & SD & $\Delta$ vs.\ latin & $2\times$ pooled SD & Verdict \\
\midrule

\texttt{no\_S} (sliding)
& 5.28495
& 0.02197
& $-0.00144$ ($-0.03\%$)
& 0.03340
& Null (within noise) \\

\texttt{no\_G} (differential)
& 5.28450
& 0.01256
& $-0.00189$ ($-0.04\%$)
& 0.02158
& Null (within noise) \\

\texttt{no\_F} (full)
& 5.29615
& 0.00284
& $+0.00976$ ($+0.18\%$)
& 0.01291
& Null (within noise) \\

\textbf{\texttt{no\_M} (Mamba-2 / SSM)}
& \textbf{5.39940}
& \textbf{0.00883}
& \textbf{$+0.11301$ ($+2.14\%$)}
& \textbf{0.01750}
& \textbf{Real effect} \\

\midrule

\textit{baseline} \texttt{latin}
& 5.28639
& 0.00867
& ---
& ---
& Reference (8 seeds) \\

\bottomrule
\end{tabular}

\end{table}

\subsection{Reading --- a monotone penalty in depth-confinement}

The four placement/reference arms form a \textbf{single monotone gradient}, ordered by how tightly each mechanism is confined within a depth band. 
As mechanism exposure becomes progressively more concentrated—from fully distributed across all depths (\texttt{latin}, \texttt{periodic}), 
to restricted within a contiguous band (\texttt{block}), to largely absent except for a single repeated type (\texttt{homo\_F})—validation loss increases monotonically. 
Table~\ref{tab:depth-confinement} summarizes this progression and illustrates that the primary penalty is associated 
with depth-wise concentration rather than any particular permutation of mechanisms.

\begin{table}[t]
\centering
\small
\caption{
Monotonic relationship between depth-wise mechanism confinement and validation-loss degradation.
Arrangements are ordered by the extent to which individual mechanisms are restricted to particular depth ranges.
Performance deteriorates as mechanism identity becomes increasingly concentrated rather than distributed throughout the network.
}
\label{tab:depth-confinement}

\begin{tabular}{lll}
\toprule
Arrangement & Mechanism distribution across depth & vs.\ \texttt{latin} \\
\midrule

\texttt{latin}, \texttt{periodic}
& Present at \textbf{all} depths (fully distributed)
& 0 (tied) \\

\texttt{block}
& Confined to a \textbf{four-layer band}
& +0.59\% \\

\texttt{homo\_F}
& Three of four mechanisms \textbf{absent}; one repeated everywhere
& +1.68\% \\

\bottomrule
\end{tabular}
\end{table}

\textbf{(i) Permutation is free.} \texttt{latin} and \texttt{periodic} differ only in column structure; their gap (0.008) is a third of the decision threshold. 
Re-permuting a distributed schedule does nothing measurable --- \textbf{the specific arrangement is not where quality comes from.}

\textbf{(ii) Distribution matters.} The \textit{only} change from \texttt{latin}/\texttt{periodic} to \texttt{block} is 
that each mechanism is now confined to a contiguous band instead of spread across depth. That alone costs \textbf{0.59\%} 
and clears the pre-registered threshold at 2.5 SD. Keeping every mechanism \textit{present throughout the depth} is worth a real, measurable amount.

\textbf{(iii) Composition matters most, and one family carries it.} Collapsing to a single type --- the limit of depth-confinement --- costs \textbf{1.68\%} (\texttt{homo\_F}). 
The leave-one-out then locates \textit{where} that composition value lives: removing \textbf{Mamba-2 (the linear-recurrent / state-space family)} costs \textbf{2.14\%} 
and is the \textbf{only} significant single removal, 
while dropping any one of the three attention variants (\texttt{full}, \texttt{sliding}, \texttt{differential}) stays within noise ($\leq$0.18\%). 
So the heterogeneity that matters is not spread evenly across the mechanisms --- it is carried almost entirely by the \textbf{linear-recurrent / SSM-family mechanism}, 
and beyond keeping that family present the attention variants are, at this scale and 1K context, largely interchangeable.

This connects directly to the flagship, and resolves an apparent tension. 
The three attention variants the ablation finds individually non-critical --- \texttt{full}, \texttt{sliding}, \texttt{differential} --- are exactly three of the flagship's own mechanisms; 
and the flagship supplies the load-bearing \textit{family} through its Mamba-style \texttt{linear} mixer (§3.1). 
So the flagship \textbf{includes}, rather than omits, the mechanism family this ablation identifies as essential. 
Whether the flagship's specific \texttt{linear} implementation carries the same weight as the proxy's Mamba-2 is the natural next check, stated in §10. 
(Note the 1.68\% and 2.14\% are two distinct measurements --- depth-confinement of the full multiset vs. a leave-one-out on a different multiset --- not points on one scale.)

\textbf{What this says about the Latin square.} The data rewards a \textit{property} --- every mechanism present at every depth, balanced --- 
and is indifferent to which permutation delivers it. The Latin square is the construction that delivers that property \textbf{provably, without search, at any depth}, 
and that \textbf{cannot} fall into the \texttt{block} failure mode by construction. Its value is therefore not a loss-curve win over \texttt{periodic}; 
it is that it is the \textit{guaranteed} generator of the one thing the loss curve actually cares about.

\subsection{A methodological note we owe the reader}

An earlier \textbf{two-seed} run showed \texttt{latin} and \texttt{periodic} cleanly separated, and we briefly treated that as a placement effect --- 
the result we \textit{wanted}, because a placement win would have been the more marketable story. 
Extending to four seeds dissolved it; extending to \textbf{eight} confirmed the dissolution ($\Delta$ 0.008, a third of the threshold). 
Had we stopped at two seeds we would have published a placement effect that does not exist.

We defended against exactly this in two ways, and recommend both: a standing rule of \textbf{no claim below four seeds}, 
and a \textbf{pre-registered decision rule}(\texttt{|\(\Delta\)| > 2\(\cdot\)pooled\_SD}, fixed 2026-07-22, before the final seeds ran) so that the threshold could not be moved to fit the outcome. 
The same rule that killed our preferred placement result is the one that certifies the \texttt{block} and \texttt{homo\_F} effects as real; 
a rule that only ever confirms what you hoped for is not a rule.

\subsection{Does the dissociation survive scale?}

A natural concern is that the preceding ablation is conducted at relatively modest scale (700.9M parameters) 
and therefore may not generalize to larger models. 
To test this possibility, we re-ran the three decisive configurations---\texttt{latin}, \texttt{homo\_F}, and \texttt{no\_M}---at \textbf{1.514B parameters}, 
representing a \textbf{$2.16\times$ increase in model size}. All other experimental factors were held constant, including the $4\times4$ Latin-square topology, 
the 1,500-step training protocol, and the token budget. Each arm was trained with three random seeds. 
Table~\ref{tab:scale-results} compares the resulting held-out cross-entropy values across the two scales.

\begin{table}[t]
\centering
\small
\caption{
Composition-axis results at two model scales (mean held-out cross-entropy).
The same experimental protocol is repeated at 700.9M and 1.514B parameters.
The homogeneous-stack penalty and the penalty associated with removing the
SSM-family mechanism both increase with scale, indicating that the composition
effect is not a small-model artifact.
}
\label{tab:scale-results}

\begin{tabular}{lcccc}
\toprule
Arm & CE (700.9M) & $\Delta$ vs.\ \texttt{latin} & CE (1.514B) & $\Delta$ vs.\ \texttt{latin} \\
\midrule

\texttt{latin}
& 5.28639
& ---
& \textbf{5.26963}
& --- \\

\texttt{homo\_F}
(homogeneous)
& 5.37540
& +1.68\%
& \textbf{5.40807}
& \textbf{+2.63\%} \\

\texttt{no\_M}
(drop SSM)
& 5.39940
& +2.14\%
& \textbf{5.43837}
& \textbf{+3.20\%} \\

\bottomrule
\end{tabular}
\end{table}

Both effects not only survive the scale increase, they \textbf{grow}. At 1.514B the homogeneous penalty is \textbf{2.63\%} 
and the cost of removing the state-space mechanism is \textbf{3.20\%} --- each larger than at pilot scale, and each 6--10$\times$ the pre-registered 
threshold (the largest seed spread among these arms is 0.019; both $|\Delta|$ are 0.14--0.17). The gap widens for a legible reason: 
\texttt{latin} \textit{improves} with scale (5.28639 $\rightarrow$ 5.26963) while \texttt{homo\_F} \textit{degrades} (5.37540 $\rightarrow$ 5.40807), 
so the heterogeneous stack converts added capacity into lower loss more effectively than the homogeneous one.

We are precise about coverage. We re-ran the \textbf{composition} arms at 1.5B, not the placement arms (\texttt{periodic}, \texttt{block}), 
so what we show to be scale-robust is the composition half of the dissociation --- that composition dominates and one mechanism carries it. 
Whether the placement null (\texttt{latin} $\approx$ \texttt{periodic}) and the clustering penalty (\texttt{block}) reproduce at 1.5B is left open. 
Nevertheless, the primary concern for any pilot study ---
whether the main finding survives scale ---
is directly addressed here. The effect not only persists at \textbf{1.514B parameters} but becomes stronger.

\section{Evaluation}

The primary objective of this work is architectural rather than leaderboard-oriented. 
The central question is not whether a Latin-square heterogeneous stack outperforms the strongest available language models, 
but whether the proposed construction preserves useful language-modeling capabilities while providing the compositional 
and placement properties studied in §6. We therefore evaluate the released model from three complementary perspectives. 
All benchmark evaluations are conducted using \texttt{lm-evaluation-harness} v0.4.11 in the 0-shot setting with $n=600$ evaluation examples per task. 
First, we measure general language understanding on established English benchmarks. 
Second, because the training corpus contains substantial Korean data, we evaluate Korean-language capability using KoBEST. 
Third, we examine whether the model captures linguistic structure rather than mere token statistics through a simple word-order perturbation analysis. 
Together, these evaluations establish that the proposed architecture remains a functional language model 
while the primary contribution of the paper—the controlled study of composition and placement effects—remains the focus of the experimental analysis.

\subsection{English}
Table~\ref{tab:english-eval} reports 0-shot English benchmark results obtained using \texttt{lm-evaluation-harness} v0.4.11. 
The purpose of this evaluation is not to establish state-of-the-art performance, 
but to verify that a heterogeneous sequence-mixer architecture retains competent language-modeling capabilities across a diverse set of commonsense reasoning, 
scientific question answering, and multiple-choice understanding tasks. 
The results indicate that the proposed architecture remains a functional general-purpose language model despite its focus on architectural experimentation rather than benchmark optimization.

\begin{table}[t]
\centering
\small
\caption{
English benchmark evaluation using \texttt{lm-evaluation-harness} v0.4.11 in the 0-shot setting.
Results are reported as task accuracy (\texttt{acc}) and normalized accuracy (\texttt{acc\_norm}) when available.
}
\label{tab:english-eval}

\begin{tabular}{lll}
\toprule
Task & acc & acc\_norm \\
\midrule
\textbf{SciQ} & \textbf{73.7} & 63.0 \\
\textbf{PIQA} & \textbf{66.3} & 65.8 \\
BoolQ & 54.3 & --- \\
\textbf{ARC-Easy} & \textbf{52.5} & 48.7 \\
WinoGrande & 51.8 & --- \\
HellaSwag & 37.2 & 41.0 \\
OpenBookQA & 20.0 & 32.6 \\
ARC-Challenge & 22.2 & 25.8 \\
\bottomrule
\end{tabular}
\end{table}

Performance is strongest on SciQ, PIQA, and ARC-Easy, suggesting that the model acquires useful commonsense and factual reasoning capabilities despite its relatively modest scale. 
More difficult benchmarks such as ARC-Challenge and OpenBookQA remain substantially below the strongest contemporary language models, 
which is expected given that the primary objective of this work is to study architectural composition rather than maximize benchmark performance.

\subsection{Korean (KoBEST)}

Table~\ref{tab:kobest} reports Korean-language evaluation results on KoBEST benchmarks. 
Because the training corpus contains dedicated Korean web and synthetic data, 
these experiments assess whether the heterogeneous sequence-mixer architecture acquires meaningful Korean-language capability rather than concentrating its capacity on English alone. 
The selected tasks cover commonsense reasoning, sentiment understanding, lexical semantics, and question answering, providing a broad view of Korean-language performance.

\begin{table}[t]
\centering
\small
\caption{
Korean-language evaluation on KoBEST benchmarks.
Results are reported as mean task scores with standard deviations across evaluation samples.
}
\label{tab:kobest}

\begin{tabular}{ll}
\toprule
Task & Score \\
\midrule
\textbf{HellaSwag} (acc\_norm) & \textbf{44.6 $\pm$ 2.2} \\
\textbf{COPA} & \textbf{57.2 $\pm$ 2.0} \\
\textbf{SentiNeg} & \textbf{55.7 $\pm$ 2.5} \\
WiC & 48.8 $\pm$ 2.0 \\
BoolQ & 47.8 $\pm$ 2.0 \\
\bottomrule
\end{tabular}
\end{table}

The strongest results are observed on COPA and SentiNeg, suggesting that the model acquires useful causal and sentiment-related reasoning capabilities in Korean. 
Performance on HellaSwag is also competitive, indicating an ability to model commonsense continuations beyond simple lexical matching. 
In contrast, WiC and BoolQ remain closer to chance-level performance, implying that fine-grained semantic discrimination and question-answering ability are less developed. 
Overall, the results demonstrate that the architecture successfully transfers a meaningful fraction of its modeling capacity to Korean 
while remaining primarily optimized for architectural investigation rather than benchmark maximization.

\subsection{Language modeling --- evidence of learned structure}

Table~\ref{tab:structure} evaluates whether the model captures linguistic structure rather than merely memorizing token statistics. 
While benchmark scores measure downstream task performance, they do not directly reveal whether a language model has learned syntactic and compositional regularities. 
We therefore compare perplexity on a natural Korean sentence, a word-order-scrambled version of the same sentence, 
and a random-token baseline. Because the token inventory remains largely unchanged between the first two conditions, 
a substantial increase in perplexity would indicate sensitivity to sentence structure rather than to token frequencies alone.

\begin{table}[t]
\centering
\small
\caption{
Perplexity under progressively disrupted linguistic structure.
The three inputs contain comparable lexical content but differ substantially in grammatical and compositional organization.
}
\label{tab:structure}

\begin{tabular}{ll}
\toprule
Input & Perplexity \\
\midrule
Natural Korean sentence & \textbf{5.4} \\
Same sentence, word order scrambled & \textbf{41.3} \\
Random tokens & 2233.4 \\
\bottomrule
\end{tabular}
\end{table}

The results reveal a strong dependence on linguistic structure. Scrambling the word order increases perplexity from 5.4 to 41.3, 
an approximately eight-fold degradation despite preserving most of the lexical content. 
Random tokens further increase perplexity by more than two orders of magnitude, indicating 
that the model assigns extremely low probability to sequences that violate both syntactic and statistical regularities. 
An $8\times$ perplexity penalty for scrambling word order while holding the token multiset fixed suggests 
that the model represents Korean \textbf{structure}, not merely token statistics. While this experiment is intentionally simple, 
it provides qualitative evidence that the heterogeneous architecture captures grammatical organization in addition to lexical co-occurrence patterns.

Because the periodic arm attains strict sliding-window balance and the Latin arm does not,
the near-equality of their validation loss (0.16\%) is direct evidence that strict local-window
balance is not what drives performance in this setting. The dissociation we report is therefore
between distributed heterogeneous composition and concentration in depth, not between two
grades of local uniformity.

\section{Validation and Scaling}

\subsection{Causal Safety}
A heterogeneous stack mixes operators along the time axis, which is exactly where causality violations hide. 
Before any quality claim can be attributed to architecture, the stack must be shown not to leak future information.

Using the audit of the companion paper, all \textbf{49 layers} pass under a negative control, 
and --- critically --- under a \textbf{positive control on the same loaded checkpoint} (16/16 injected faults localized exactly, across NSA, hybrid, 
and linear-attention layers). The verdict is unchanged after loading trained weights.

We report the positive control because a clean verdict without one is not evidence; see the companion paper for the case where that distinction mattered.

\subsection{Scaling the Construction}
\texttt{Aether-6B-11Attn-base} places \textbf{eleven} mechanisms --- attention, Mamba-2, Hyena, GDN, MLA among them --- 
on an \textbf{11$\times$11 Latin square over 121 layers}. 
It is a mid-training research artifact, released as-is.

Its purpose here is narrow: to show the construction is not specific to $N=7$. 
Whether the composition findings of §6 hold at $N=11$ is open.

\section{Discussion and Conclusion}

\subsection{Limitations}
\begin{enumerate}
\item \textbf{The ablation is a pilot proxy.} 700.9M--1.514B parameters, 1,500 steps, \textbf{four mechanisms over sixteen layers} --- not the 6.59B, seven-mechanism, 49-layer flagship. 
It tests the \textit{construction principle}, not the shipped configuration. The composition axis is confirmed to hold --- and strengthen --- at 1.514B (§6.5), 
but the placement axis (\texttt{periodic}, \texttt{block}) was not re-run at that scale, and neither axis was tested at $N=7$ or at flagship scale.
\item \textbf{The permutation null is scoped to \textit{distributed} schedules.} We find no measurable difference \textit{among balanced, 
distributed arrangements} (\texttt{latin} vs \texttt{periodic}); we explicitly do \textbf{not} claim placement never matters --- \texttt{block} shows it does when distribution is broken. 
"Permutation is free" means free \textit{within the distributed regime}, not everywhere.
\item \textbf{Benchmark scores are not competitive with the strongest models of similar size.} What we offer is openness and a controlled ablation, not leaderboard position.
\item \textbf{Mechanism labels are coarse.} Seven labels map onto five distinct implementations.
\item \textbf{Single training run per configuration at 6.59B.} The ablation has seeds; the flagship does not.
\item \textbf{No KV cache in the released architecture}, so generation is slow. This is a property of the release, documented rather than hidden.
\end{enumerate}

\subsection{Conclusion}
Every layer using the same attention was never a finding; it was a habit. When we tested it --- parameter count, placement, and mechanism set each varied one at a time, 
eight seeds per arm, under a pre-registered rule --- the answer came out as a single monotone gradient: 
quality degrades in lockstep with \textbf{how tightly each mechanism is confined to a depth band}. 
Spread across all depths, the permutation is free. Confined to a band, you pay 0.59\%. Collapsed to one type, you pay 1.68\%.

So the operative property is \textit{balance and distribution across depth} --- and the specific arrangement that achieves it is immaterial to the loss. 
That is what makes the \textbf{Latin square} a principled construction rather than an incidental one: it is a search-free, depth-invariant, provable generator of that property, 
and it \textbf{cannot, by construction, fall into the concentration failure mode}. A plain periodic cycle shares those virtues and ties it in the data; 
we adopt the Latin square as the \textit{canonical, maximally-uniform} member of that family, 
generalizing to any prime \textit{N}. Placement-invariance is not a null result to hide --- it is the evidence that the property, not the permutation, 
is the lever, and therefore the reason a provable construction is the right default rather than a searched one.

We release the model, the data recipe, the code, the logs, and the intermediate checkpoints so that every number above can be contradicted.

\bibliographystyle{unsrt}   
\bibliography{references}   

\newpage

\section*{Appendix A --- Artifacts}

A central goal of this project is reproducibility. Beyond releasing model weights, we provide the artifacts required to reconstruct, inspect, 
and verify the claims made throughout the paper. Table~\ref{tab:artifacts} summarizes the publicly released resources, 
including model checkpoints, architecture implementations, training recipes, evaluation code, and intermediate training states.

The release is intended to support multiple forms of verification. Researchers can reproduce the reported experiments, inspect architectural decisions, 
analyze the training trajectory through intermediate checkpoints, and independently evaluate both the proposed Latin-square construction and the accompanying ablation results. 
Together, these artifacts aim to reduce the gap between reported results and independently verifiable evidence.

\begin{table}[t] 
\centering
\small 
\caption{ Artifacts released with Aether, including model checkpoints, source code, training recipes, and evaluation resources. } 
\label{tab:artifacts}
\begin{tabular}{ll}
\toprule
Item & Location \\
\midrule
Base model & \texttt{FINAL-Bench/Aether-7B-5Attn} \\
Instruction-tuned & \texttt{FINAL-Bench/Aether-7B-5Attn-it} \\
Second checkpoint, same architecture & \texttt{FINAL-Bench/AETHER-7B-7Attn-base} \\
11-mechanism extension & \texttt{FINAL-Bench/Aether-6B-11Attn-base} \\
Ablation code & \texttt{pilot\_model.py}, \texttt{pilot\_train.py} \\
Causal audit & companion paper, Appendix A \\
\bottomrule
\end{tabular}
\end{table}

Several items deserve special attention. Intermediate checkpoints allow the learning dynamics of the model to be studied rather than only the final trained state. 
Complete training logs provide a direct record of optimization behavior and enable independent verification of reported trends. 
Combined with the released training-data recipe and architecture source code, 
these resources make it possible to reproduce substantial portions of the experimental pipeline without relying on undocumented implementation details.

\end{document}